\documentclass[10pt,twocolumn,letterpaper]{article}

\usepackage{cvpr}
\usepackage{times}
\usepackage{epsfig}
\usepackage{graphicx}
\usepackage{amsmath}
\usepackage{amssymb}
\usepackage{eso-pic}
\usepackage{helvet}
\usepackage{courier}
\usepackage{url}
\usepackage{graphicx}
\usepackage{mathrsfs}
\usepackage{multirow}
\usepackage{subfigure}
\usepackage{comment}
\usepackage{marvosym}
\usepackage{bbm}
\usepackage{enumerate}
\usepackage[dvipsnames]{xcolor}
\usepackage{xcolor}
\usepackage{soul}
\usepackage[utf8]{inputenc}
\usepackage[small]{caption}

\usepackage{booktabs}
\usepackage{amsfonts}
\usepackage{nicefrac}
\usepackage{microtype}
\usepackage[numbers]{natbib}
\usepackage[breaklinks=true,bookmarks=false]{hyperref}

\cvprfinalcopy
\def\cvprPaperID{5165}

\ifcvprfinal\fi

\begin{document}

\title{Memory In Memory: A Predictive Neural Network for Learning Higher-Order Non-Stationarity from Spatiotemporal Dynamics}

\author{
Yunbo Wang\thanks{Equal contribution, in alphabetical order} , Jianjin Zhang$^{*}$, Hongyu Zhu, {Mingsheng Long} (\Letter), Jianmin Wang, and Philip S. Yu\\
KLiss, MOE; BNRist; School of Software, Tsinghua University, China\\
Research Center for Big Data, Tsinghua University, China\\
Beijing Key Laboratory for Industrial Big Data System and Application\\
{\small\tt \{wangyb15,zhang-jj16\}@mails.tsinghua.edu.cn, \{mingsheng,jimwang\}@tsinghua.edu.cn}
}

\maketitle
\thispagestyle{empty}

\begin{abstract}
   Natural spatiotemporal processes can be highly non-stationary in many ways, e.g. the low-level non-stationarity such as spatial correlations or temporal dependencies of local pixel values; and the high-level variations such as the accumulation, deformation or dissipation of radar echoes in precipitation forecasting. From Cram\'{e}r's Decomposition \cite{cramer1961some}, any non-stationary process can be decomposed into deterministic, time-variant polynomials, plus a zero-mean stochastic term. By applying differencing operations appropriately, we may turn time-variant polynomials into a constant, making the deterministic component predictable. 
   
   However, most previous recurrent neural networks for spatiotemporal prediction do not use the differential signals effectively, and their relatively simple state transition functions prevent them from learning too complicated variations in spacetime. We propose the \textbf{Memory In Memory} (MIM) networks and corresponding recurrent blocks for this purpose. The MIM blocks exploit the differential signals between adjacent recurrent states to model the non-stationary and approximately stationary properties in spatiotemporal dynamics with two cascaded, self-renewed memory modules. By stacking multiple MIM blocks, we could potentially handle higher-order non-stationarity. The MIM networks achieve the state-of-the-art results on four spatiotemporal prediction tasks across both synthetic and real-world datasets. We believe that the general idea of this work can be potentially applied to other time-series forecasting tasks.
\end{abstract}

\section{Introduction}
\label{sec:Introduction}

\begin{figure}[tbh]
  \centering
  \includegraphics[width=\columnwidth]{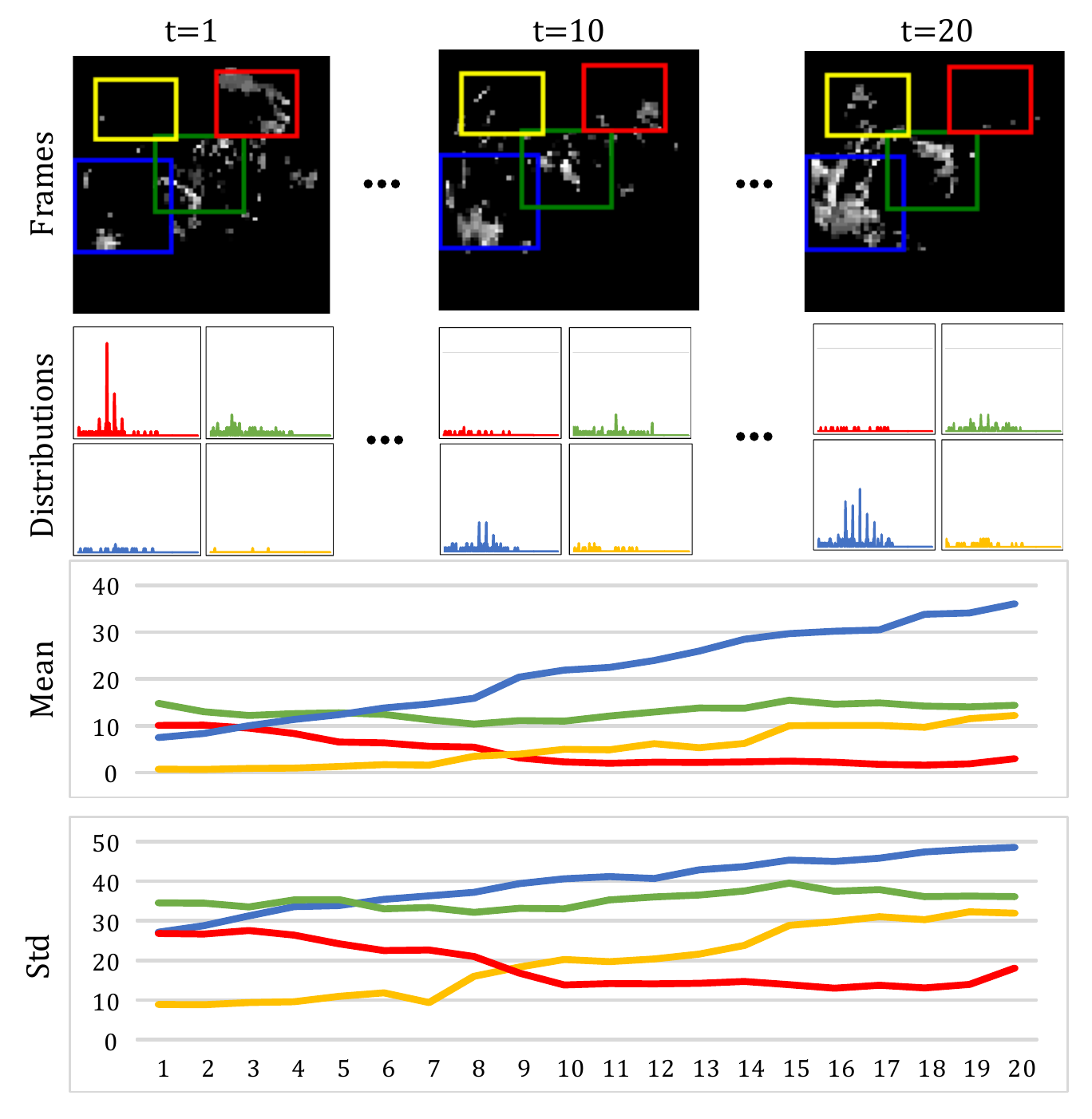}
  \caption{An example of $20$ consecutive radar maps to illustrate the complicated non-stationarity in precipitation forecasting. 
  \textbf{First row:} radar maps, in which the whiter pixels show higher precipitation probability. 
  \textbf{Second, third, last row:} pixel values' distributions, means and standard deviations for corresponding local regions that are identified by bounding boxes of different colors.
  Note that different regions have different variation trends, making the spatiotemporal prediction problem extremely challenging.} 
  \label{fig:radar_intro}
\end{figure}

Natural spatiotemporal processes exhibit complex non-stationarity in both space and time, where neighboring pixels exhibit local dependencies, and their joint distributions are changing over time. 
Learning higher-order properties underlying the spatiotemporal non-stationarity is particularly significant for many video prediction tasks. 
Examples include modeling highly complicated real-world systems such as traffic flows \citep{zhang2017deep,xu2018predcnn} and weather conditions \citep{shi2015convolutional,wang2017predrnn}. 
A well-performed predictive model is expected to learn the intrinsic variations in consecutive spatiotemporal context, which can be seen as a combination of the stationary component and the deterministic non-stationary component.

A great challenge in non-stationary spatiotemporal prediction is how to effectively capture higher-order trends regarding each pixel and its local area.
For example, when making precipitation forecasting, one should carefully consider the complicated and diverse local trends on the evolving radar maps, shown as Figure \ref{fig:radar_intro}. But this problem is extremely difficult due to the complicated non-stationarity in both space and time.
Most prior work handles trend-like non-stationarity with recursions of CNNs \citep{zhang2017deep,xu2018predcnn} or relatively simple state transitions in RNNs \citep{shi2015convolutional,wang2017predrnn}. 
The lack of non-stationary modeling capability prevents reasoning about uncertainties in spatiotemporal dynamics and partially leads to the blurry effect of the predicted frames. 
%

We attempt to resolve this problem by proposing a generic RNNs architecture that is more effective in non-stationarity modeling. 
We find that though the forget gates in the recurrent predictive models could deliver, select, and discard information in the process of memory state transitions, they are too simple to capture higher-order non-stationary trends in high-dimensional time series. 
In particular, the forget gates in the recent PredRNN model \citep{wang2017predrnn} does not work appropriately on precipitation forecasting: about $80\%$ of them are saturated over all timestamps, implying almost time-invariant memory state transitions. 
In other words, future frames are predicted by approximately linear extrapolations.

In this paper, we focus on improving the memory transition functions of RNNs.  
Most statistical forecasting methods in classic time series analysis assume that the non-stationary trends can be rendered approximately stationary by performing suitable transformations such as \textit{differencing}.
We introduce this idea to RNNs and propose a new RNNs building block named \textit{Memory In Memory} (MIM), which leverages the differential information between neighboring hidden states in the recurrent paths. 
MIM can be viewed as an improved version of LSTM~\cite{Hochreiter1997Long}, whose forget gate is replaced by another two embedded long short-term memories.

MIM has the following characteristics: \textbf{(1)} It creates unified modeling for the spatiotemporal non-stationarity by differencing neighboring hidden states rather than raw images.
\textbf{(2)} By stacking multiple MIM blocks, our model has a chance to gradually stationarize the spatiotemporal process and make it more predictable.
\textbf{(3)} Note that over-differencing is no good for time series prediction, as it may inevitably lead to a loss of information. This is another reason that we apply differencing in memory transitions rather than all recurrent signals, e.g. the input gate and the input modulation gate.
\textbf{(4)} MIM has one memory cell adopted from LSTMs as well as two additional recurrent modules with their own memories embedded in the transition path of the first memory. We use these modules to respectively model the higher-order non-stationary and approximately stationary components of the spatiotemporal dynamics.
The proposed MIM networks achieve the state-of-the-art results on multiple prediction tasks, including a widely used synthetic dataset and three real-world datasets.

\section{Related Work}
\label{sec:Related}

\subsection{ARIMA Models for Time Series Forecasting}

Our model is inspired by the Autoregressive Integrated Moving Average (ARIMA) models.
A time-series random variable whose power spectrum remains constant over time can be viewed as a combination of signal and noise.  An ARIMA model aims to separate the signal from the noise. The obtained signal is then extrapolated into the future.
In theory, it tackles time series forecasting by transforming the non-stationary process to stationary through differencing \citep{box2015time}.

\begin{figure*}[t]
  \centering
  \includegraphics[width=0.8\textwidth]{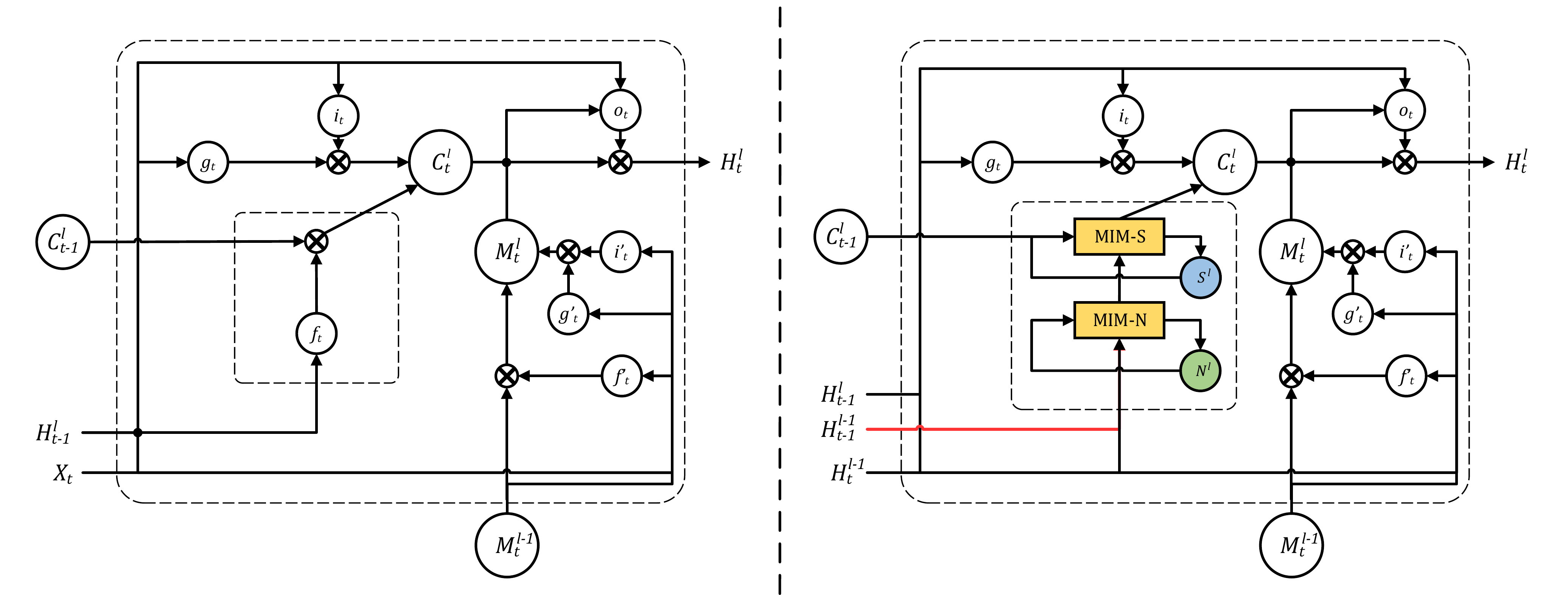}
  \vspace{-5pt}
  \caption{The ST-LSTM block \citep{wang2017predrnn} in the left plot and the proposed Memory In Memory (MIM) block in the right plot. MIM is designed to introduce two recurrent modules (yellow squares) to replace the forget gate (dashed box) in ST-LSTM. MIM-N is the non-stationary module and MIM-S is the stationary module. Note that the MIM block cannot be used in the first layer so the input $\mathcal{X}_t$ is replaced by $\mathcal{H}_t^{l-1}$.
  }
  \label{fig:node}
  \vspace{-5pt}
\end{figure*}

\subsection{Deterministic Spatiotemporal Prediction}

Spatiotemporal non-stationary processes are more complicated, as the joint distribution of neighboring pixel values is varying in both space and time. 
Like low-dimensional time series, they can also be decomposed into deterministic and stochastic components.
Recent work in neural networks explored spatiotemporal prediction from these two aspects.

CNNs \citep{krizhevsky2012imagenet} and RNNs \citep{Sutskever2014Sequence} have been widely used for learning the deterministic spatial correlations and temporal dependencies from videos.
Ranzato \textit{et al.} \citep{Ranzato2014Video} defined a recurrent model predicting frames in a discrete space of patch clusters. 
Srivastava \textit{et al.} \citep{srivastava2015unsupervised} introduced the sequence to sequence LSTM network from language modeling to video prediction. But this model can only capture temporal variations. 
To learn spatial and temporal variations in a unified network structure, Shi \textit{et al.} \citep{shi2015convolutional} integrated the convolution operator into recurrent state transition functions, and proposed the Convolutional LSTM. 
Finn \textit{et al.} \citep{Finn2016Unsupervised} developed an action-conditioned video prediction model that can be further used in robotics planning when combined with the model predictive control methods.
Villegas \textit{et al.} \citep{Villegas2017Decomposing} and Patraucean \textit{et al.} \citep{patraucean2015spatio} presented recurrent models based on the convolutional LSTM that leverage optical flow guided features. 
Kalchbrenner \textit{et al.} \citep{Kalchbrenner2016Video} proposed the Video Pixel Network (VPN) that encodes the time, space, color structures of videos as a four-dimensional dependency chain. 
It achieves sharp prediction results but suffers from a high computational complexity.
Wang \textit{et al.} \citep{wang2017predrnn,wang2018predrnn++} extended the convolutional LSTM with zigzag memory flows, which provides a great modeling capability for short-term video dynamics. 
Adversarial learning \citep{Goodfellow2014Generative,Denton2015Deep} has been increasingly used in video generation or prediction \citep{Mathieu2015Deep,vondrick2016generating,denton2017unsupervised,tulyakov2017mocogan,wichers2018hierarchical}, as it aims to solve the multi-modal training difficulty of the future prediction and helps generate less blurry frames.

However, the high-order non-stationarity of video dynamics has not been thoroughly considered by the above work, whose temporal transition methods are relatively simple, either controlled by the recurrent gate structures or implemented by the recursion of the feed-forward network.
By contrast, our model is characterized by exploiting high-order differencing to mitigate the non-stationary learning difficulty.

\subsection{Stochastic Spatiotemporal Prediction}

Some recent methods \citep{xue2016visual,denton2018stochastic,lee2018stochastic} attempted to model the stochastic component of video dynamics using Variational Autoencoder \citep{kingma2013auto}.
%
%
These methods increase the prediction diversity but are difficult to evaluate and require to run a great number of times for a satisfactory result. 
In this paper, we focus on the deterministic part of spatiotemporal non-stationarity. 
More specifically, this work attempts to stationarize the complicated spatiotemporal processes and make their deterministic components in the future more predictable by proposing new RNNs architecture for non-stationarity.

\section{Methods}
\label{sec:Methods}

As mentioned above, the spatiotemporal non-stationarity remains under-explored and its differential features have not been fully exploited by previous methods using neural networks.
In this section, we first present the \textit{Memory In Memory} (MIM) blocks for learning about the higher-order non-stationarity from RNNs memory transitions.
We then discuss a new RNN architecture, which interlinks multiple MIM blocks with diagonal state connections, for modeling the differential information in the spatiotemporal prediction.
By stacking multiple MIM blocks, we could potentially learn higher-order non-stationarity from spatiotemporal dynamics.
The proposed MIM state transition approach can be integrated into all LSTM-like units.
We choose the Spatiotemporal LSTM (ST-LSTM) \citep{wang2017predrnn} as our base network for a trade-off between prediction accuracy and computation simplicity.
ST-LSTM is characterized by a dual-memory structure, $\mathcal{C}_t^l$ and $\mathcal{M}_t^l$, as shown in Figure \ref{fig:node} (left). 
The corresponding zigzag memory flow of $\mathcal{M}_t^l$, as illustrated by the black arrows in Figure \ref{fig:structure}, strengthens its short-term modeling capability by increasing the recurrent transition depth.
%
%

\subsection{Memory In Memory Blocks}

We observe that the complex dynamics in spatiotemporal sequences can be handled more effectively as a combination of stationary variations and non-stationary variations. 
Suppose we have a video sequence showing a person walking at a constant speed. The velocity can be seen as a stationary variable and the swing of the legs should be considered as a non-stationary process, which is apparently more difficult to predict. 
Unfortunately, the forget gate in previous LSTM-like models is a simple gating structure that struggles to capture the non-stationary variations in spacetime. 
In preliminary experiments, we find that the majority of forget gates in the recent PredRNN model \citep{wang2017predrnn} are saturated, implying that the units always remember stationary variations.

\begin{figure}[t]
  \centering
  \includegraphics[width=\columnwidth]{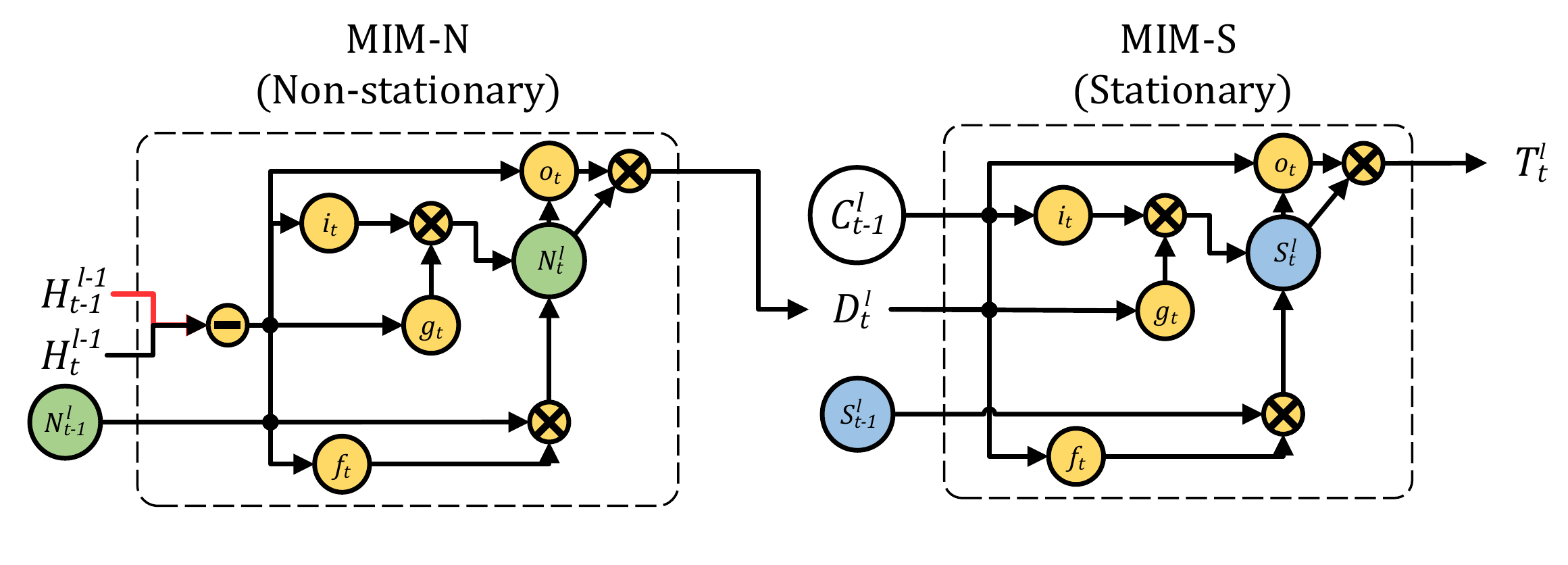}
  \vspace{-20pt}
  \caption{The non-stationary module (MIM-N) and the stationary module (MIM-S), which are interlinked in a cascaded structure in the MIM block. Non-stationarity is modeled by differencing.}
  \label{fig:subunits}
\end{figure}

The Memory In Memory (MIM) block is enlightened by the idea of modeling the non-stationary variations using a series of cascaded memory transitions instead of the simple, saturation-prone forget gate in ST-LSTM. 
As compared in Figure \ref{fig:node} (the smaller dashed boxes), two cascaded temporal memory recurrent modules are designed to replace the temporal forget gate $f_t$ in ST-LSTM. The first module additionally taking $\mathcal{H}_{t-1}^{l-1}$ as input is used to capture the non-stationary variations based on the differencing $(\mathcal{H}_{t}^{l-1} - \mathcal{H}_{t-1}^{l-1})$ between two consecutive hidden representations. So we name it the \textbf{non-stationary module} (shown as \textbf{MIM-N} in Figure \ref{fig:subunits}). 
It generates differential features $\mathcal{D}_t^l$ based on the difference-stationary assumption \citep{percival1993spectral}. 
The other recurrent module takes as inputs the output $\mathcal{D}_t^l$ of the MIM-N module and the outer temporal memory $\mathcal{C}_{t-1}^{l}$ to capture the approximately stationary variations in spatiotemporal sequences. So we call it the \textbf{stationary module} (shown as \textbf{MIM-S} in Figure \ref{fig:subunits}). 
By replacing the forget gate with the final output $\mathcal{T}_t^l$ of the cascaded non-stationary and stationary modules (as shown in Figure~\ref{fig:node}), the non-stationary dynamics can be captured more effectively. Key calculations inside a \textbf{MIM} block can be shown as follows:
\begin{equation}\label{equ:MIM}
  \small
  \begin{split}
  & g_t = \tanh(W_{xg} \ast \mathcal{H}_t^{l-1} + W_{hg} \ast \mathcal{H}_{t-1}^l + b_g) \\
  & i_t = \sigma(W_{xi} \ast \mathcal{H}_t^{l-1} + W_{hi} \ast \mathcal{H}_{t-1}^l + b_i) \\
  & \mathcal{D}_t^l = \text{MIM-N}(\mathcal{H}_t^{l-1}, \mathcal{H}_{t-1}^{l-1}, \mathcal{N}_{t-1}^l) \\
  & \mathcal{T}_t^l = \text{MIM-S}(\mathcal{D}_t^l, \mathcal{C}_{t-1}^l, \mathcal{S}_{t-1}^l) \\
  & \mathcal{C}_t^l = \mathcal{T}_t^l + i_t \odot g_t \\
  & g_t^\prime = \tanh(W_{xg}^\prime \ast \mathcal{H}_t^{l-1} + W_{mg} \ast \mathcal{M}_t^{l-1} + b_g^\prime) \\
  & i_t^\prime = \sigma(W_{xi}^\prime \ast \mathcal{H}_t^{l-1} + W_{mi} \ast \mathcal{M}_t^{l-1} + b_i^\prime) \\
  & f_t^\prime = \sigma(W_{xf}^\prime \ast \mathcal{H}_t^{l-1} + W_{mf} \ast \mathcal{M}_t^{l-1} + b_f^\prime) \\
  & \mathcal{M}_t^l = f_t^\prime \odot \mathcal{M}_t^{l-1} + i_t^\prime \odot g_t^\prime \\
  & o_t = \sigma(W_{xo} \ast \mathcal{H}_t^{l-1} + W_{ho} \ast \mathcal{H}_{t-1}^l + W_{co} \ast \mathcal{C}_t^l + W_{mo} \ast \mathcal{M}_t^l + b_o) \\ 
  & \mathcal{H}_t^l = o_t \odot \tanh(W_{1\times1} \ast [\mathcal{C}_t^l, \mathcal{M}_t^l]),\\
  \end{split}
  \normalsize
\end{equation}
where $\mathcal{S}$ and $\mathcal{N}$ denote the horizontally-transited memory cells in the non-stationary module (MIM-N) and stationary module (MIM-S) respectively; $\mathcal{D}$ is the differential features learned by MIM-N and fed into MIM-S; $\mathcal{T}$ is the memory passing the virtual ``forget gate''; and $\ast$ denotes convolution.

The cascaded structure enables end-to-end modeling of different orders of non-stationary dynamics.
It is based on the difference-stationary assumption that differencing a non-stationary process repeatedly will likely lead to a stationary one \citep{percival1993spectral}.
A schematic of MIM-N and MIM-S is presented in Figure \ref{fig:subunits}. We present the detailed calculations of \textbf{MIM-N} as follows:
\begin{equation}\label{equ:non-stationary}
  \small
  \begin{split}
  & g_t = \tanh(W_{xg} \ast (\mathcal{H}_t^{l-1} - \mathcal{H}_{t-1}^{l-1}) + W_{ng} \ast \mathcal{N}_{t-1}^l + b_g) \\
  & i_t = \sigma(W_{xi} \ast (\mathcal{H}_t^{l-1} - \mathcal{H}_{t-1}^{l-1}) + W_{ni} \ast \mathcal{N}_{t-1}^l + b_i)\\
  & f_t = \sigma(W_{xf} \ast (\mathcal{H}_t^{l-1} - \mathcal{H}_{t-1}^{l-1}) + W_{nf} \ast \mathcal{N}_{t-1}^l + b_f) \\
  & \mathcal{N}_t^l = f_t \odot \mathcal{N}_{t-1}^l + i_t \odot g_t \\
  & o_t = \sigma(W_{xo} \ast (\mathcal{H}_t^{l-1} - \mathcal{H}_{t-1}^{l-1}) + W_{no} \ast \mathcal{N}_t^l + b_o) \\ 
  & \mathcal{D}_t^l =  \text{MIM-N}(\mathcal{H}_t^{l-1}, \mathcal{H}_{t-1}^{l-1}, \mathcal{N}_{t-1}^l) = o_t \odot \tanh(\mathcal{N}_t^l),\\
  \end{split}
\end{equation}
where all gates $g_t$, $i_t$, $f_t$ and $o_t$ are updated by incorporating the frame difference $(\mathcal{H}_t^{l-1} - \mathcal{H}_{t-1}^{l-1})$, which highlights the non-stationary variations in the spatiotemporal sequence.
The detailed calculations of \textbf{MIM-S} are shown as follows:
\begin{equation}\label{equ:stationary}
  \small
  \begin{split}
  & g_t = \tanh(W_{dg} \ast \mathcal{D}_t^l + W_{cg} \ast \mathcal{C}_{t-1}^l + b_g) \\
  & i_t = \sigma(W_{di} \ast \mathcal{D}_t^l + W_{ci} \ast \mathcal{C}_{t-1}^l + b_i)\\
  & f_t = \sigma(W_{df} \ast \mathcal{D}_t^l + W_{cf} \ast \mathcal{C}_{t-1}^l + b_f) \\
  & \mathcal{S}_t^l = f_t \odot \mathcal{S}_{t-1}^l + i_t \odot g_t \\
  & o_t = \sigma(W_{do} \ast \mathcal{D}_t^l + W_{co} \ast \mathcal{C}_{t-1}^l + W_{so} \ast \mathcal{S}_t^l + b_o) \\ 
  & \mathcal{T}_t^l = \text{MIM-S}(\mathcal{D}_t^l, \mathcal{C}_{t-1}^l, \mathcal{S}_{t-1}^l) = o_t \odot \tanh(\mathcal{S}_t^l),\\
  \end{split}
\end{equation}
which takes the memory cells $\mathcal{C}_{t-1}^l$ and the differential features $\mathcal{D}_t^l$ generated by MIM-N as input. As can be validated, the stationary module provides a gating mechanism to adaptively decide whether to trust the original memory $\mathcal{C}_{t-1}^l$ or the differential features $\mathcal{D}_t^l$. 
If the differential features vanish, indicating that the non-stationary dynamics is not prominent, then MIM-S will mainly reuse the original memory.
Otherwise, if the differential features are prominent, then MIM-S will overwrite the original memory and focus more on the non-stationary dynamics.

\subsection{Memory In Memory Networks}

\textbf{Stacking multiple MIM blocks}, our model has a chance to capture \textbf{higher orders of non-stationarity}, gradually stationarizes the spatiotemporal process and makes the future sequence more predictable.
The key idea of this architecture is to deliver necessary hidden states for generating differential features and best facilitating non-stationarity modeling.

\begin{figure}[t]
  \centering
  \includegraphics[width=\columnwidth]{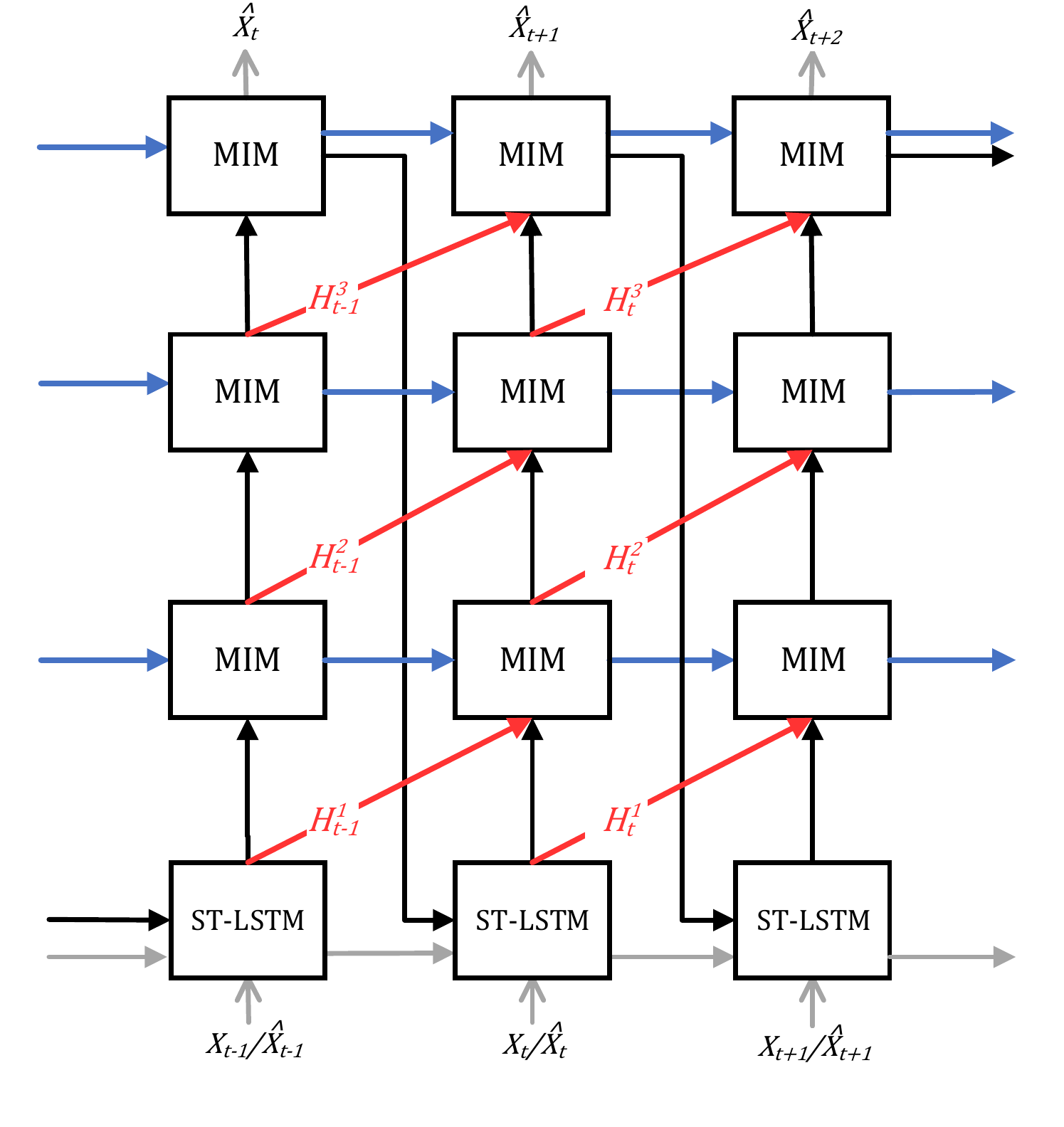}
  \vspace{-20pt}
  \caption{A MIM network with three MIMs and one ST-LSTM. 
  \textbf{\textcolor{red}{Red arrows:}} the \textit{diagonal} state transition paths of $\mathcal{H}$ for differential modeling. 
  \textbf{\textcolor{RoyalBlue}{Blue arrows:}} the \textit{horizontal} transition paths of the  memory cells $\mathcal{C}$, $\mathcal{N}$ and $\mathcal{S}$.
  \textbf{Black arrows:} the \textit{zigzag} state transition paths of $\mathcal{M}$.
  \textbf{Input:} the input can be either the ground truth frame for input sequence, or the generated frame at previous timestamp.
  \textbf{Output:} one frame is generated at each timestamp.}
  \label{fig:structure}
\end{figure}

A schematic of our proposed diagonal recurrent architecture is shown in Figure \ref{fig:structure}. 
We deliver the hidden states $\mathcal{H}_{t-1}^{l-1}$ and $\mathcal{H}_t^{l-1}$ to the \textit{Memory In Memory} (MIM) block at timestamp $t \neq 1$ and layer $l \neq 1$ to generate the differenced features for further use. 
These connections are shown as diagonal arrows in Figure \ref{fig:structure}. As the first layer does not have any previous layer, we simply use the Spatiotemporal LSTM (ST-LSTM) \citep{wang2017predrnn} to generate its hidden presentations.
Note that, the temporal differencing is performed by subtracting hidden state $\mathcal{H}_{t}^{l-1}$ from the hidden state $\mathcal{H}_{t-1}^{l-1}$ in MIM. 
Compared to differencing neighboring raw images directly, differencing temporally adjacent hidden states can reveal the non-stationarity more evidently, as the spatiotemporal variations in local areas have been encoded into the hidden representations through the bottom ST-LSTM layer.

Another distinctive feature of the MIM networks resides in the horizontal state transition paths.
As the MIM blocks have two cascaded temporal memory modules to capture the non-stationary and stationary dynamics respectively, we further deliver the two temporal memories (denoted by $\mathcal{N}$ for the  \textit{non-stationary} memory and by $\mathcal{S}$ for the \textit{stationary} memory) along the blue arrows in Figure \ref{fig:structure}.

The MIM networks generate one frame at one timestamp. Calculations of the entire model with one ST-LSTM and $(L-1)$ MIMs can be presented as follows (for $2\leq l\leq L$). Note that there is no MIM block that is marked as $\text{MIM}_1$.
\begin{equation}\label{equ:MIM_structrue}
  \small
  \begin{split}
  \mathcal{H}_t^1,\mathcal{C}_t^1,\mathcal{M}_t^1 = \text{ST-LSTM}_1(& \mathcal{X}_t, \mathcal{H}_{t-1}^1, \mathcal{C}_{t-1}^1, \mathcal{M}_{t-1}^L) \\
  \mathcal{H}_t^l,\mathcal{C}_t^l,\mathcal{M}_t^l,\mathcal{N}_t^{l},\mathcal{S}_t^{l} = \text{MIM}_l(& \mathcal{H}_t^{l-1}, \mathcal{H}_{t-1}^l, \mathcal{C}_{t-1}^l, \mathcal{M}_t^{l-1},\\
  & 
  \mathcal{N}_{t-1}^{l},\mathcal{S}_{t-1}^{l}). \\
  \end{split}
\end{equation}

We formulate high-order non-stationarity as high-order polynomials based on Cram\'{e}r's Decomposition \cite{cramer1961some}. In the space-time contexts, it refers to the varying trends of statistics of pixel values.
The order of non-stationary polynomials can be reduced by a couple of differencing operations. We blend this idea from time-series analysis with deep learning: stacking MIMs with differential inputs layer by layer.

\section{Experiments}
\label{sec:Experiments}
In this section, we evaluate the proposed MIM model using four datasets for spatiotemporal prediction: a synthetic dataset with moving digits, a real traffic flow dataset, a real radar echo dataset, and a human action dataset.
Here are some common settings all over these datasets.
Our model has four layers in all experiments, including one ST-LSTM layer as the first layer and three MIMs.
The number of feature channels in each MIM block is $64$, as a trade-off of prediction accuracy and memory efficiency. 
All models are trained with the $\ell_2$ loss, using the ADAM optimizer \citep{Kingma2014Adam} with a learning rate of $0.001$. The mini-batch size is set to $8$.
We apply the layer normalization \citep{Ba2016Layer} to the compared models in order to reduce the covariate shift problem \citep{ioffe2015batch}. 
Besides, we apply the scheduled sampling \citep{bengio2015scheduled} to all models to stitch the discrepancy between training and inference.
%
%
Code and models are available at \textcolor{magenta}{\url{https://github.com/Yunbo426/MIM}}.


\subsection{Moving MNIST}

The standard Moving MNIST is a synthetic dataset with grayscale image sequences of flying digits. We follow exactly the experimental settings in PredRNN \citep{wang2017predrnn}.
%
%

\begin{figure}[t]
  \centering
  \includegraphics[width=\columnwidth]{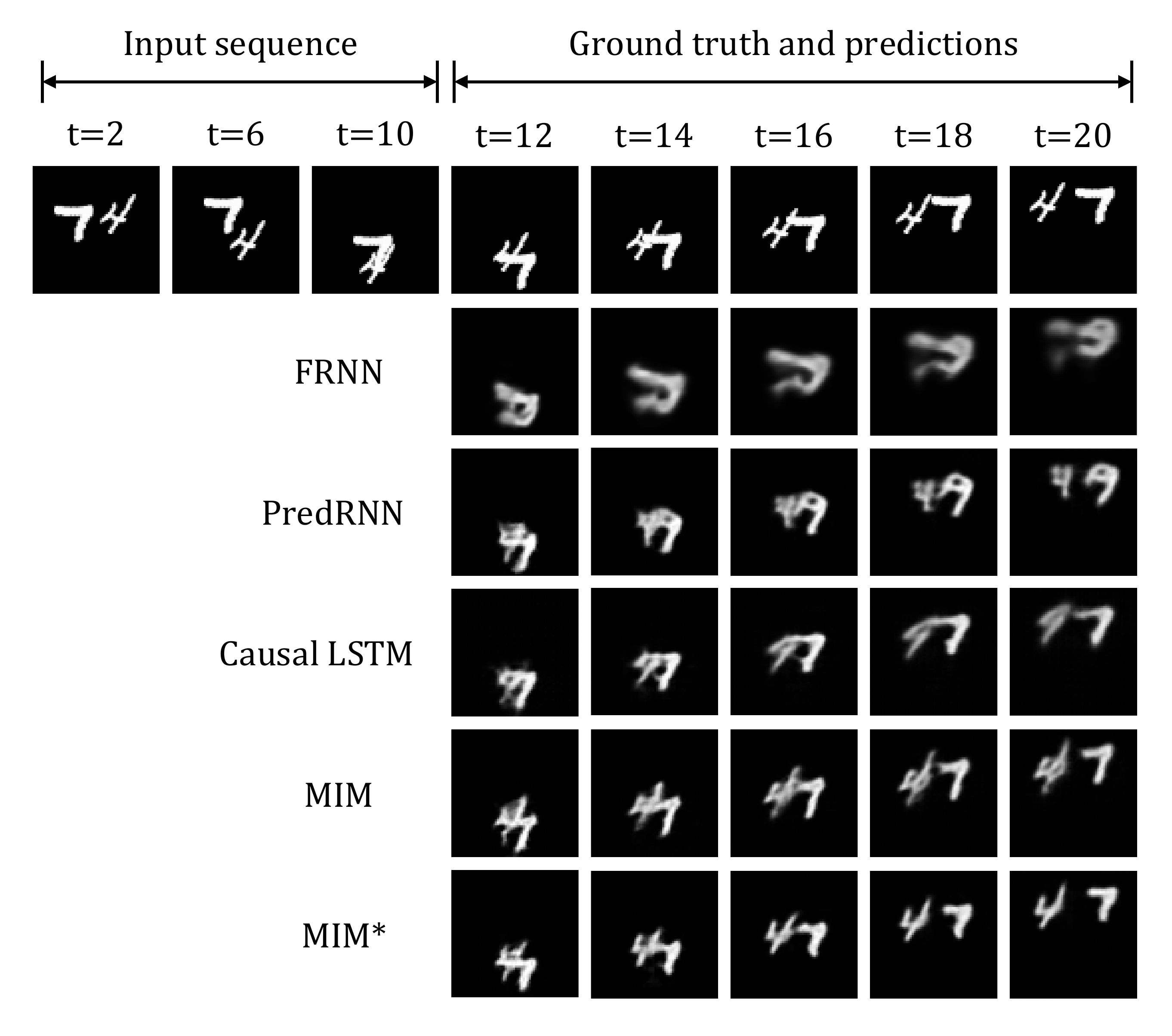}
  \vspace{-15pt}
  \caption{Prediction examples on the standard Moving MNIST. All models predict 10 frames into the future by observing 10 previous frames. The output frames are shown at two frames intervals.}
  \label{fig:mnist_results}
\end{figure}

We visualize a sequence of predicted frames on the standard Moving MNIST test set in Figure \ref{fig:mnist_results}. 
This example is challenging, as severe occlusions exist near the junction of the input sequence and the output sequence. The occlusions can be viewed as information bottleneck, in which the mean and variance of the spatiotemporal process meet drastic changes, indicating the presence of a high-order non-stationarity.
The generated images of MIM are more satisfactory, less blurry than those of other models. Actually, we cannot even tell the digits in the last frames generated by other models. We may conclude that MIM shows more capability in capturing complicated non-stationary variations. 

\begin{table}[t]
\begin{center}
\begin{tabular}{|l|ccc|}
\hline
Model & SSIM & MSE & MAE\\
\hline\hline
FC-LSTM \citep{srivastava2015unsupervised}& 0.690 & 118.3 & 209.4 \\ 
ConvLSTM \citep{shi2015convolutional} & 0.707 & 103.3 & 182.9 \\ 
TrajGRU \citep{shi2017deep} & 0.713 & 106.9 & 190.1 \\ 
CDNA \citep{Finn2016Unsupervised} & 0.721 & 97.4 & 175.3 \\ 
DFN \citep{de2016dynamic} & 0.726 & 89.0 & 172.8 \\ 
FRNN \citep{oliu2017folded} & 0.813 & 69.7 & 150.3 \\ 
VPN baseline \citep{Kalchbrenner2016Video} & 0.870 & 64.1 & 131.0 \\
PredRNN \citep{wang2017predrnn} & 0.867 & 56.8 & 126.1 \\
Causal LSTM \citep{wang2018predrnn++} & 0.898 & 46.5 & 106.8 \\
\hline
MIM & 0.874 & 52.0 & 116.5 \\
\textbf{MIM*} & \textbf{0.910} & \textbf{44.2} & \textbf{101.1} \\
\hline
\end{tabular}
\end{center}
\vspace{-10pt}
\caption{A comparison for predicting $10$ frames on Moving MNIST dataset. All models have comparable numbers of parameters. \textbf{MIM*} is network using Causal LSTM \cite{wang2018predrnn++} as the first layer, and integrating the cascaded MIM-N and MIM-S modules into the Causal LSTM memory cells. This result shows that MIM is a generic mechanism for improving recurrent memory transitions.}
\label{tab:mnist_10_result}
\end{table}

We use the per-frame structural similarity index measure (SSIM) \citep{Wang2004Image}, the mean square error (MSE) and the mean absolute error (MAE) to evaluate our models. A lower MSE or MAE, or a higher SSIM indicates a better prediction.
As shown in Table \ref{tab:mnist_10_result}, our proposed MIM model approaches the state-of-the-art results on the standard Moving MNIST dataset. 
In particular, we construct another model named MIM* by using Causal LSTM \cite{wang2018predrnn++} as the first layer, and integrating the cascaded MIM-N and MIM-S modules into the Causal LSTM memory cells, using them to replace the temporal forget gates in Causal LSTMs. 
This result shows that the \textit{memory in memory} mechanism is not specifically designed for the ST-LSTM; instead, it is a generic mechanism for improving RNNs memory transitions. 
Though in other parts of this paper, we use ST-LSTM as our base structure for a trade-off between prediction accuracy and computational complexity, we can see that MIM performs better than its ST-LSTM (PredRNN) baseline, while MIM* also performs better than its Causal LSTM baseline.
We also adopt the gradient-based sharpness metric from \citep{Mathieu2015Deep} to measure the sharpness of the generated images. As shown in Table \ref{tab:sharpness}, MIM rises the sharpness score by ${16\%}$ over PredRNN.

\begin{table}[h]
\begin{center}
\begin{tabular}{|l|c|}
\hline
Model & Gradient-Based Sharpness \\
\hline\hline
FRNN \citep{oliu2017folded} & 24.99  \\
PredRNN \citep{wang2017predrnn} & 23.29 \\
\textbf{MIM} & \textbf{27.05} \\
\hline
\end{tabular}
\end{center}
\vspace{-10pt}
\caption{Sharpness evaluation on Moving MNIST.}
\label{tab:sharpness}
\end{table}

We further testify \textbf{the necessity of cascading inner recurrent modules} by respectively removing the stationary modules or non-stationary modules.
As illustrated in Table \ref{tab:mnist_10_result_self}, the MIM network without MIM-N works slightly better than that without MIM-S.
Also, either of them has significant improvements over the PredRNN model in MSE/MAE, showing the necessity of cascading them in a unified network. When MIM-N and MIM-S are interlinked, the entire MIM model achieves the best performance.

\begin{table}[h]
\begin{center}
\begin{tabular}{|l|ccc|}
\hline
Model & SSIM & MSE & MAE\\
\hline\hline
MIM (without MIM-N) & 0.858 & 54.4 & 124.8 \\
MIM (without MIM-S) & 0.853 & 55.7 & 125.5 \\
\textbf{MIM} & \textbf{0.874} & \textbf{52.0} & \textbf{116.5} \\
\hline
\end{tabular}
\end{center}
\vspace{-10pt}
\caption{Ablation study with respect to the MIM block.}
\label{tab:mnist_10_result_self}
\end{table}

We study \textbf{the sensitivity of our model to the number of MIM blocks}.
As stacking $2$--$4$ recurrent layers is a common practice, we evaluate models with $2/3/4$ MIMs on Moving MNIST. 
The proposed model with $3$ MIMs performs best.
It is a trade-off: applying too few MIMs leads to inadequate non-stationary modeling capability while the excessively deep recurrent model leads to training difficulty.

\begin{table}[h]
\begin{center}
\begin{tabular}{|c|cc|}
\hline
\# MIM blocks & SSIM & MSE \\
\hline\hline
2 & 0.870 & 54.6 \\
\textbf{3} & \textbf{0.874} & \textbf{52.0} \\
4 & 0.859 & 53.7 \\
\hline
\end{tabular}
\end{center}
\vspace{-10pt}
\caption{The sensitivity of our model to the number of MIM blocks.}
\label{tab:mim_layer_number}
\end{table}

\begin{figure}[h]
\centering
\includegraphics[width=0.55\columnwidth]{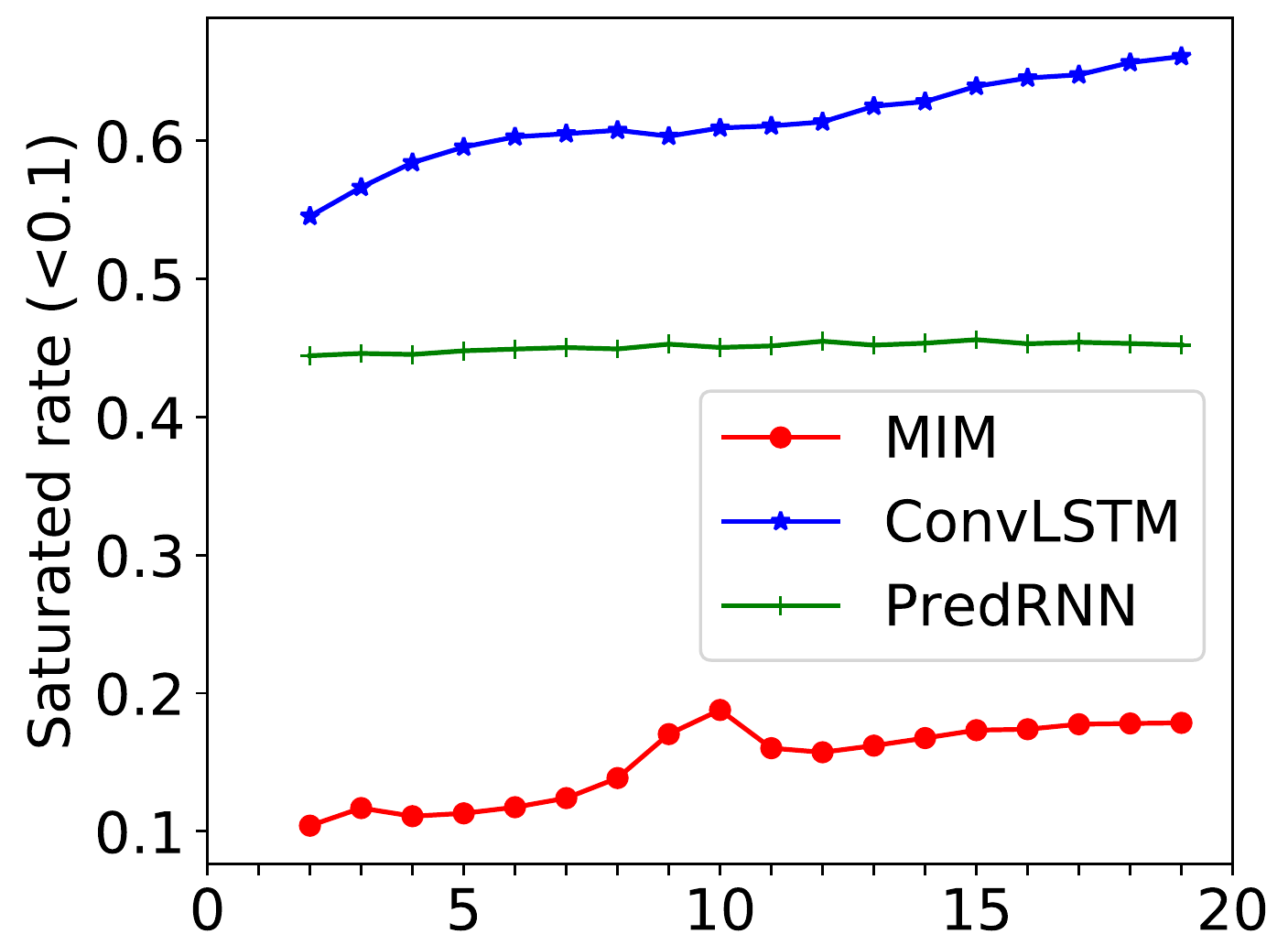}
\vspace{-5pt}
\caption{The saturated rate of forget gates that corresponds to  $|\mathcal{T}_{t}/\mathcal{C}_{t-1}|<0.1$ for MIM, and ${f}_{t}<0.1$ for other models.}
\label{fig:saturate}
\end{figure}

\textbf{The forget gates are easily saturated} in PredRNN as well as other LSTM-based spatiotemporal prediction models such as ConvLSTM \citep{shi2015convolutional}.
As shown in Figure \ref{fig:saturate}, a great portion of forget gates of these models are close to $0$, indicating that the \textit{long short-term memories} do not work properly. 
As the non-stationary signals within short-term variations are hard to be captured, these models have to refresh the memory states to convey these short-term hidden representations. 
Our model mitigates this problem and makes better use of long-term variations through the proposed MIM block. 
As MIM-N mainly reduces the non-stationarity, short-term tendencies become easier to be captured and more neurons in MIM-S can be used for handling long-term variations.

\subsection{TaxiBJ Traffic Flow}

\begin{figure}[h]
  \vspace{-5pt}
  \centering
  \includegraphics[width=\columnwidth]{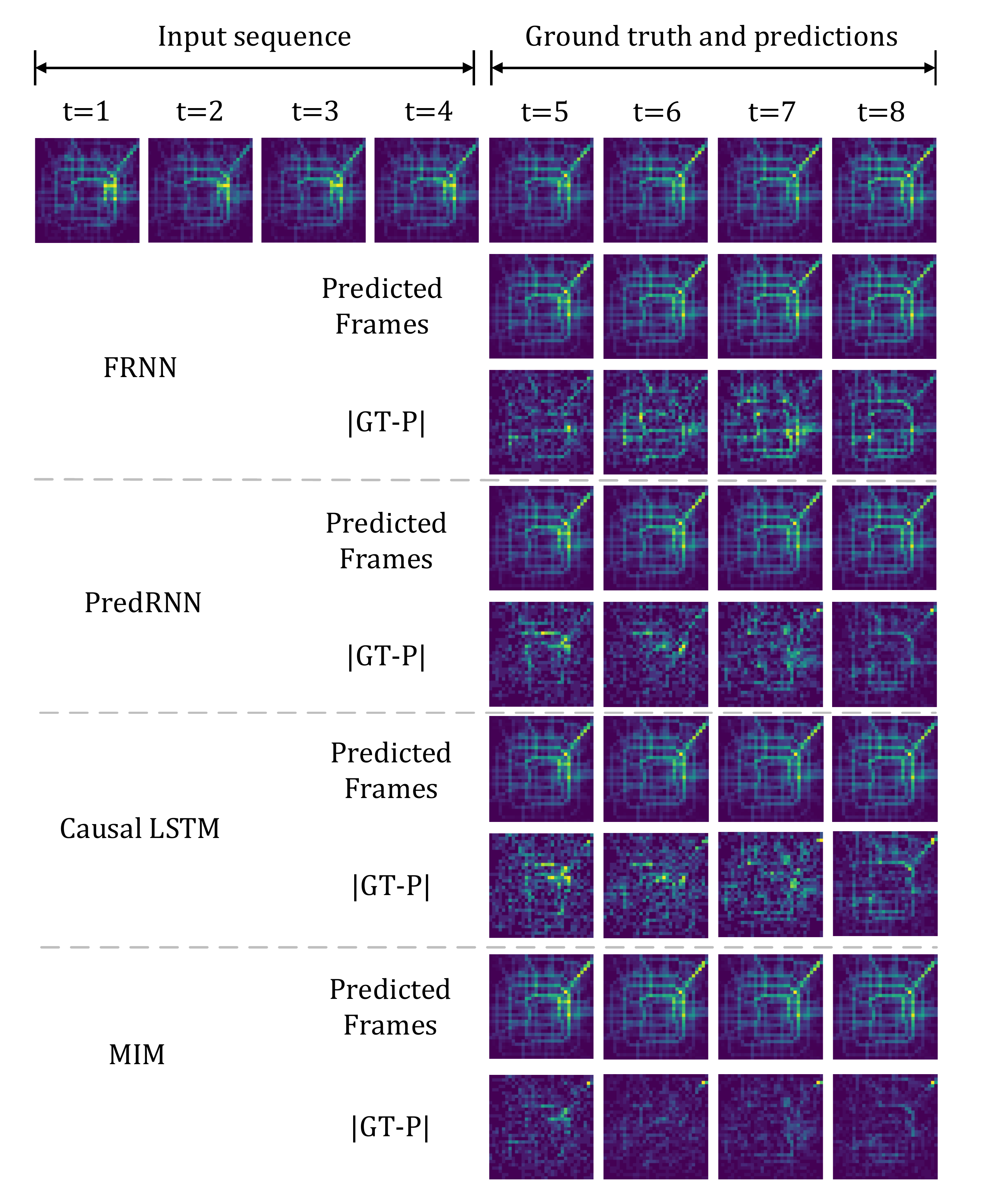}
  \vspace{-15pt}
  \caption{Prediction examples on TaxiBJ dataset. For ease of comparison, we also visualize the difference between the ground truth frames (GT) and predicted frames (P).}
  \label{fig:taxibj_results}
\end{figure}

Traffic flows are collected from the chaotic real-world environment. They will not vary uniformly over time, and there are strong temporal dependencies between the traffic conditions at neighboring timestamps.
Each frame in TaxiBJ is a $32 \times 32 \times 2$ grid of image. Two channels represent the traffic flow entering and leaving the same district. 
We normalize the data to $[0, 1]$ and follow the experimental settings of ST-ResNet \citep{zhang2017deep}, which yields the previous state-of-the-art results on this dataset. Each sequence contains $8$ consecutive frames, $4$ for the inputs and $4$ for the predictions. 
%
We show the quantitative results in Table \ref{tab:taxibj_4_result} and the qualitative results in Figure \ref{fig:taxibj_results}. To make the comparisons conspicuous, we also visualize the difference between the predictions and the ground truth images. Obviously, MIM shows the best performance in all predicted frames among all compared models, with the lowest difference intensities.

\begin{table}[h]
    \small
    \begin{center}
    \addtolength{\tabcolsep}{-1pt} 
    \begin{tabular}{|l|cccc|}
    \hline
    Model & Frame 1 & Frame 2 & Frame 3 & Frame 4\\
    \hline\hline
    ST-ResNet \citep{zhang2017deep} & 0.460 & 0.571 & 0.670 & 0.762 \\ 
    VPN \citep{Kalchbrenner2016Video} & 0.427 & 0.548 & 0.645 & 0.721 \\ 
    FRNN \citep{oliu2017folded} & 0.331 & 0.416 & 0.518 & 0.619 \\ 
    PredRNN \citep{wang2017predrnn}  & 0.318 & 0.427 & 0.516 & 0.595 \\
    Causal LSTM \citep{wang2018predrnn++}  & 0.319 & 0.399 & 0.500 & 0.573 \\
    \hline
    \textbf{MIM} & \textbf{0.309} & \textbf{0.390} & \textbf{0.475} & \textbf{0.542} \\
    \hline
    \end{tabular}
    \end{center}
    \vspace{-10pt}
    \caption{Per-frame MSE calculated with data in the range of $[0,1]$ on the TaxiBJ dataset. All compared models take $4$ historical traffic flow images as inputs, and predict the next $4$ images (traffic flows for the next two hours).}
    \label{tab:taxibj_4_result}
\end{table}

\subsection{Radar Echo}

\begin{figure}[h]
  \centering
  \includegraphics[width=\columnwidth]{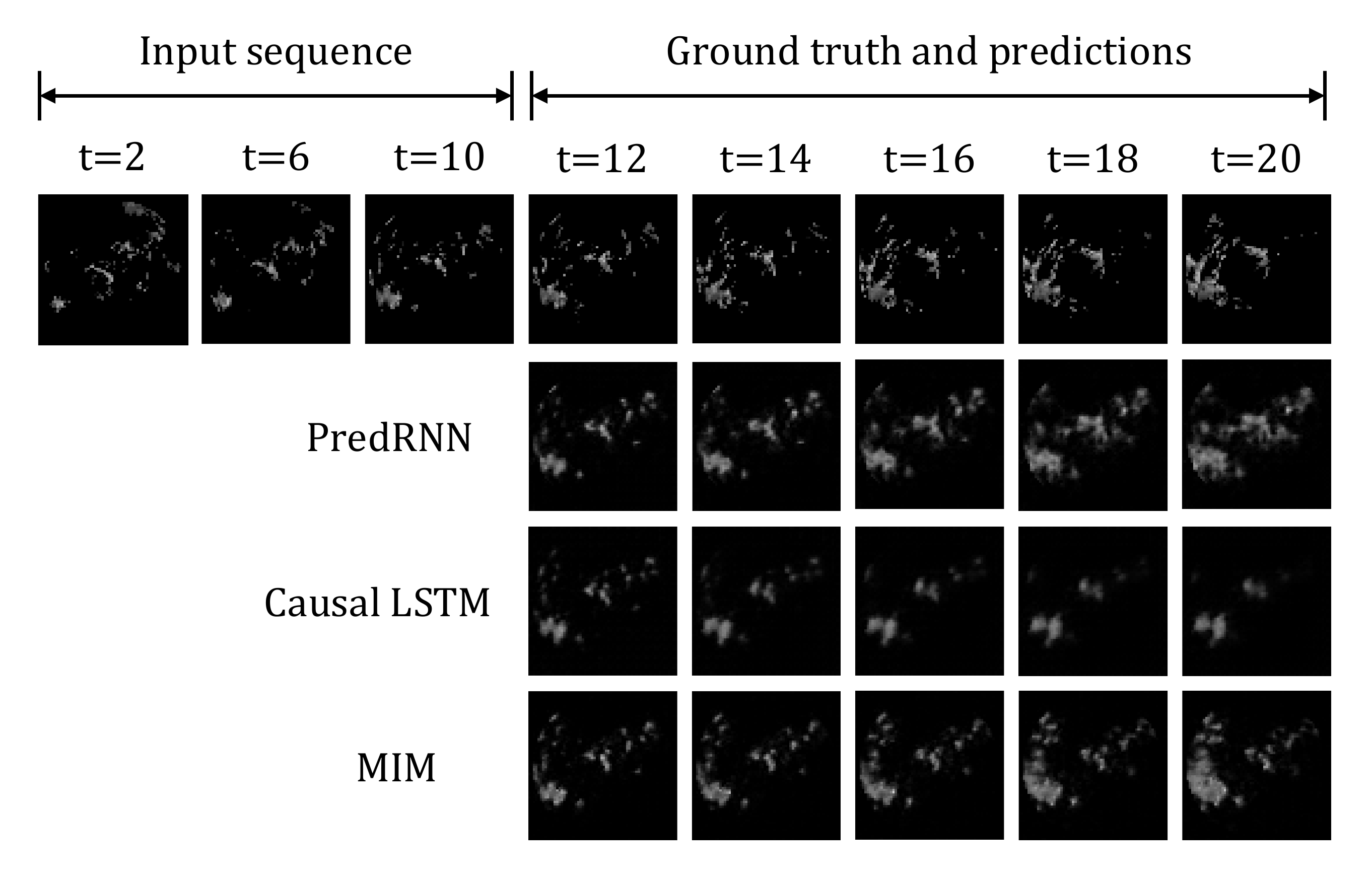}
  \vspace{-15pt}
  \caption{Examples of the next-hour predictions of radar echoes, where higher pixel values indicate higher precipitation probabilities.}
  \label{fig:radar_results}
\end{figure}

The radar echo dataset contains evolving radar maps that were collected every $6$ minutes, from May 1st, 2014 to June 30th, 2014. 
Each frame is a $64 \times 64 \times 1$ grid of image, covering $64 \times 64$ square kilometers. 
%
We predict $10$ timestamps into the future at a time interval of $6$ minutes, covering the next hour. 
We visualize the generated radar maps in Figure \ref{fig:radar_results}. 
We can see that the evolution of radar echoes is a highly non-stationary process. The accumulation, deformation, and dissipation of the radar echoes are happening at every moment.
In this showcase, the echoes in the bottom left corner aggregate while those in the upper right corner dissipate.
Only MIM captures the movement of the echoes correctly.

\begin{table}[h]
    \begin{center}
    \addtolength{\tabcolsep}{-1pt} 
    \begin{tabular}{|l|cccc|}
    \hline
    Model & MSE & CSI-30 & CSI-40 & CSI-50 \\
    \hline\hline
    FRNN \citep{oliu2017folded} & 52.5 & 0.254 & 0.203 & 0.163 \\  
    PredRNN \citep{wang2017predrnn} & 31.8 & 0.401 & 0.378 & 0.306 \\
    Causal LSTM \citep{wang2018predrnn++} & 29.8 & 0.362 & 0.331 & 0.251 \\
    \textbf{MIM} & \textbf{27.8} & \textbf{0.429} & \textbf{0.399} & \textbf{0.317}\\
    \hline
    \end{tabular}
    \end{center}
    \vspace{-10pt}
    \caption{A comparison for predicting $10$ frames on the subsets of the radar dataset. All of the models are also trained with $10$ target frames and made to predict $10$ future frames at test time.}
    \label{tab:radar_20_result}
\end{table}

\begin{figure}[t]
  \centering
  \subfigure[MSE]{
  \includegraphics[width=0.231\textwidth]{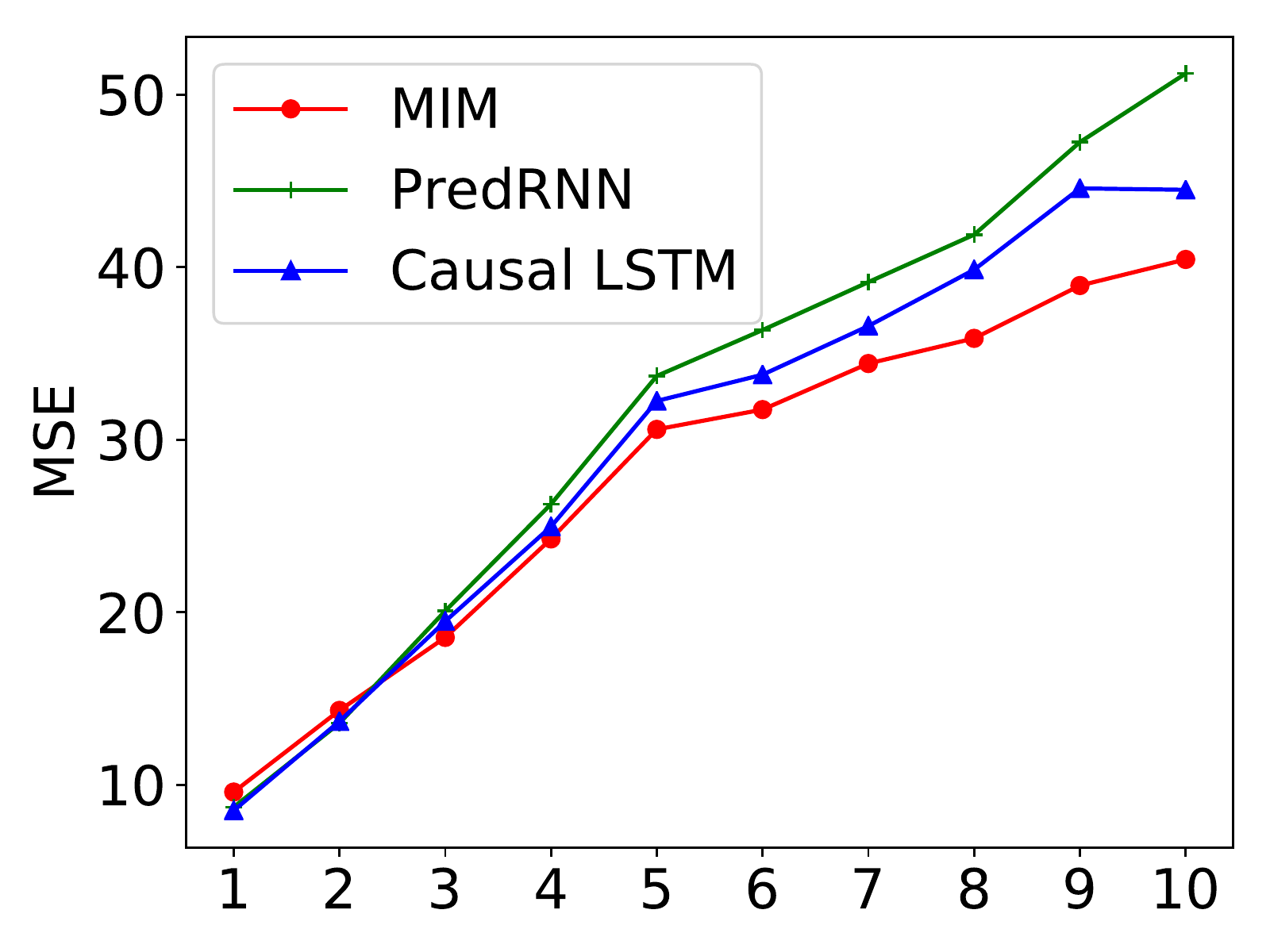}}
  \subfigure[CSI-30]{
  \includegraphics[width=0.231\textwidth]{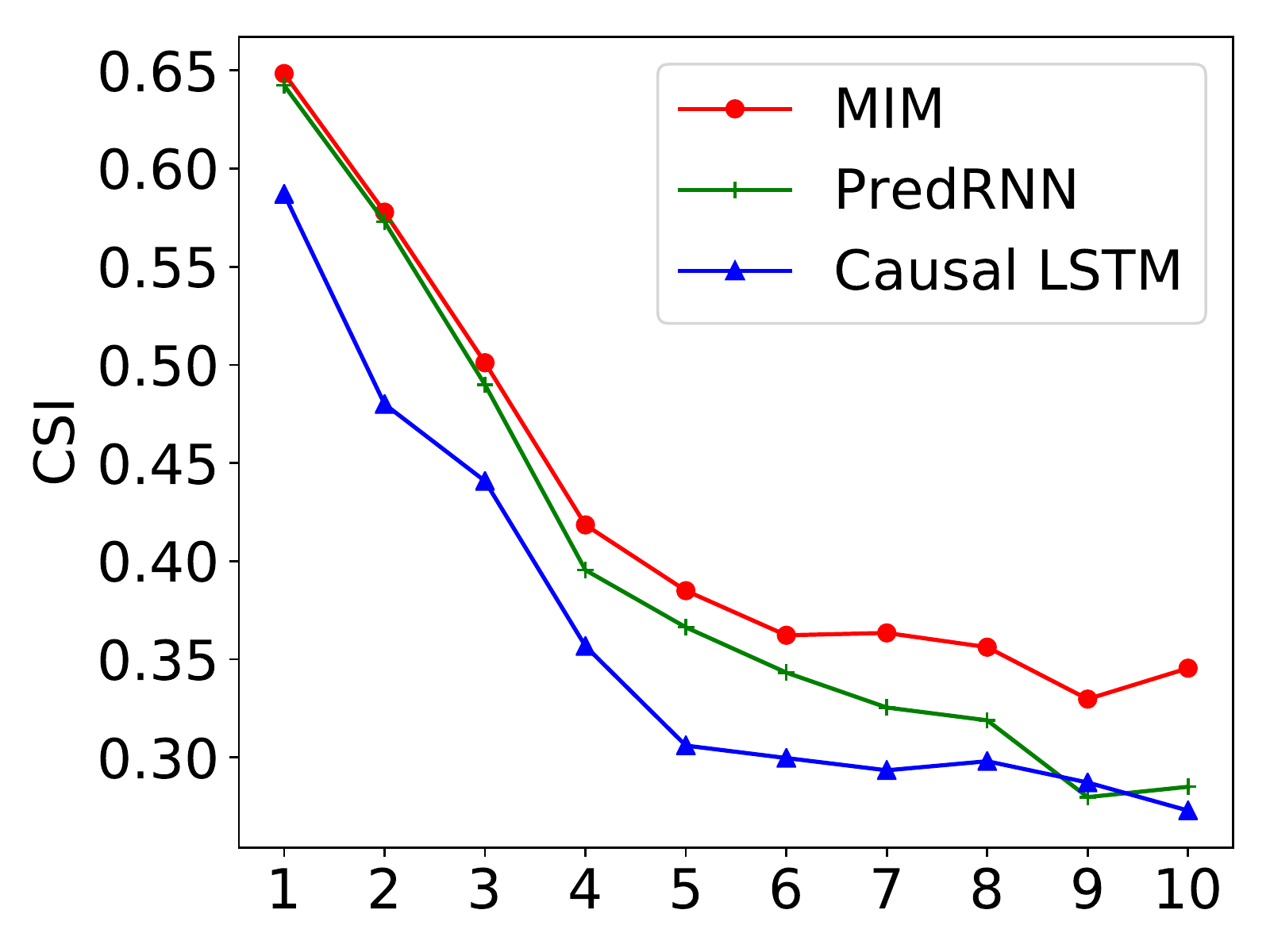}}
  \subfigure[CSI-40]{
  \includegraphics[width=0.231\textwidth]{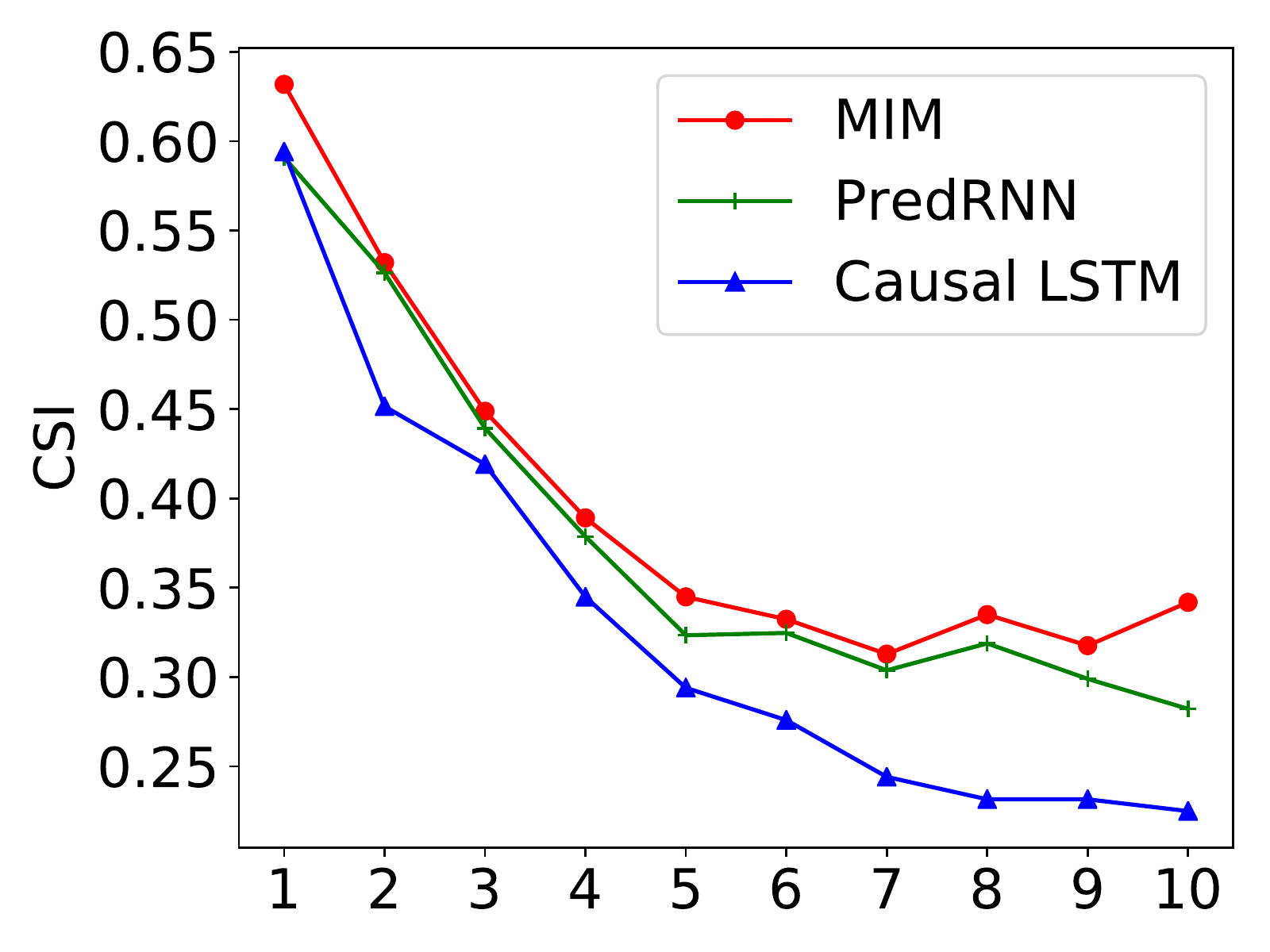}}
  \subfigure[CSI-50]{
  \includegraphics[width=0.231\textwidth]{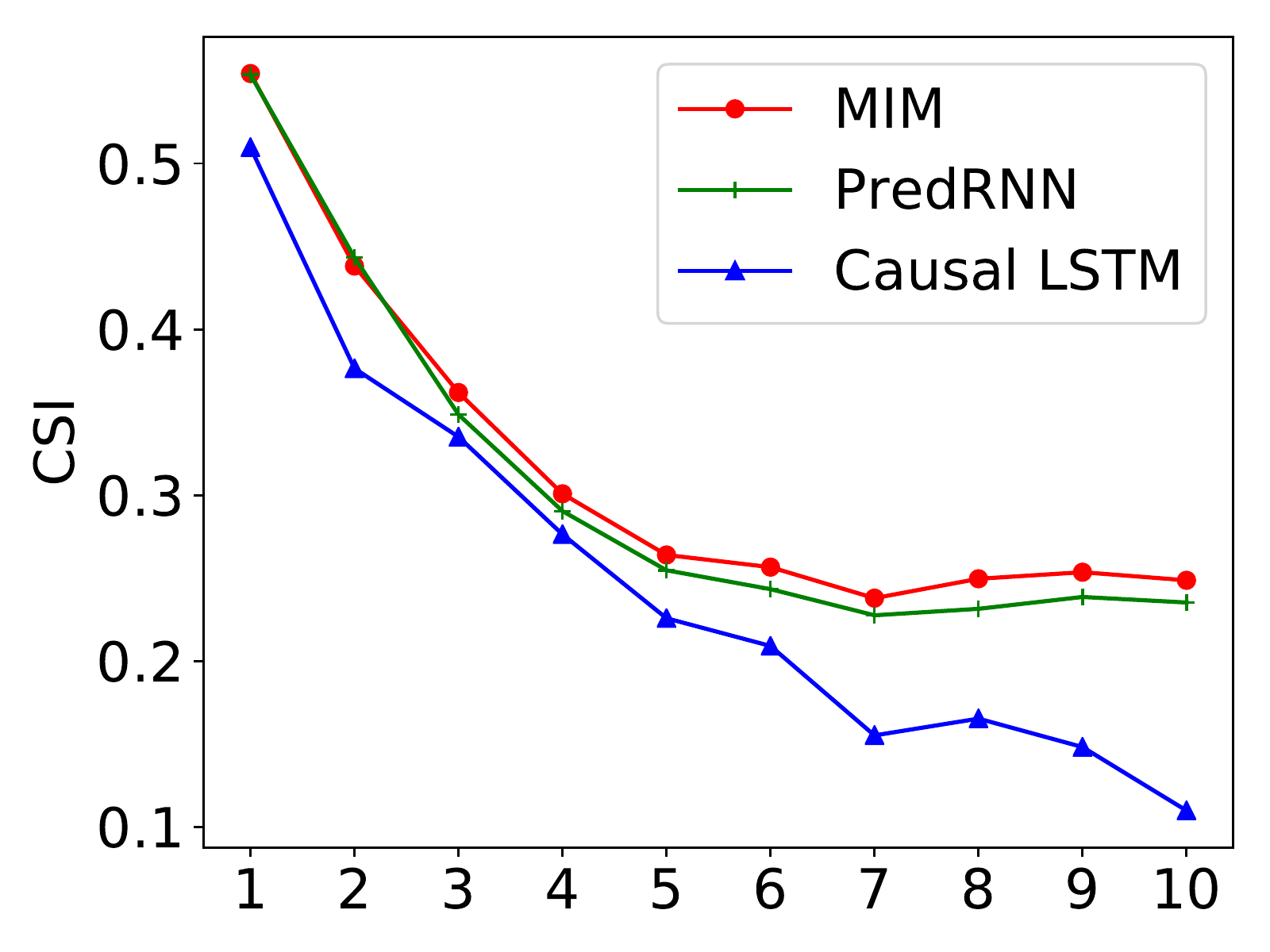}}
  \vspace{-10pt}
  \caption{Frame-wise comparisons of the next $10$ generated radar maps. Lower MSE curves or higher CSI curves indicate better forecasting results. The MIM network is the most high-performing method over all timestamps in the forecasting horizon.}
  \label{fig:radar_mse_csi}
\end{figure}

%
We evaluate the generated radar echoes by MSE in Table \ref{tab:radar_20_result}, 
and then convert pixel values to radar echo intensities in dBZ. 
We respectively choose $30$ dBZ, $40$ dBZ and $50$ dBZ as thresholds to calculate the Critical Success Index (CSI). CSI is defined as $\text{CSI}=\frac{\text{hits}}{\text{hits}+\text{misses}+\text{falsealarms}}$, where hits corresponds to \textit{true positive}, misses corresponds to \textit{false positive}, and false alarms corresponds to \textit{false negative}.
%
%
A higher CSI denotes a better prediction result. MIM consistently outperforms other models in both MSE and CSI.
Figure \ref{fig:radar_mse_csi} shows the frame-wise MSE/CSI. 
%
CSI-$40$ and CSI-$50$ indicate the probabilities of severe weather conditions. MIM performs best though predicting severe weather is non-trivial due to the long tail distributions of the pixel values.

\subsection{Human3.6M}


The Human3.6M dataset \citep{h36m_pami} contains human actions of $17$ scenarios, including $3.6$ million poses and corresponding images. We train the models using only the ``Walking'' scenario. The RGB images in Human3.6M dataset are originally $1000 \times 1000 \times 3$, and resized to $128 \times 128 \times 3$ in our experiments. 
We generate $4$ future frames given the previous $4$ RGB frames. We use subjects S1, S5, S6, S7, S8 for training, and S9, S11 for testing. 

\begin{figure}[t]
    \centering
    \includegraphics[width=\columnwidth]{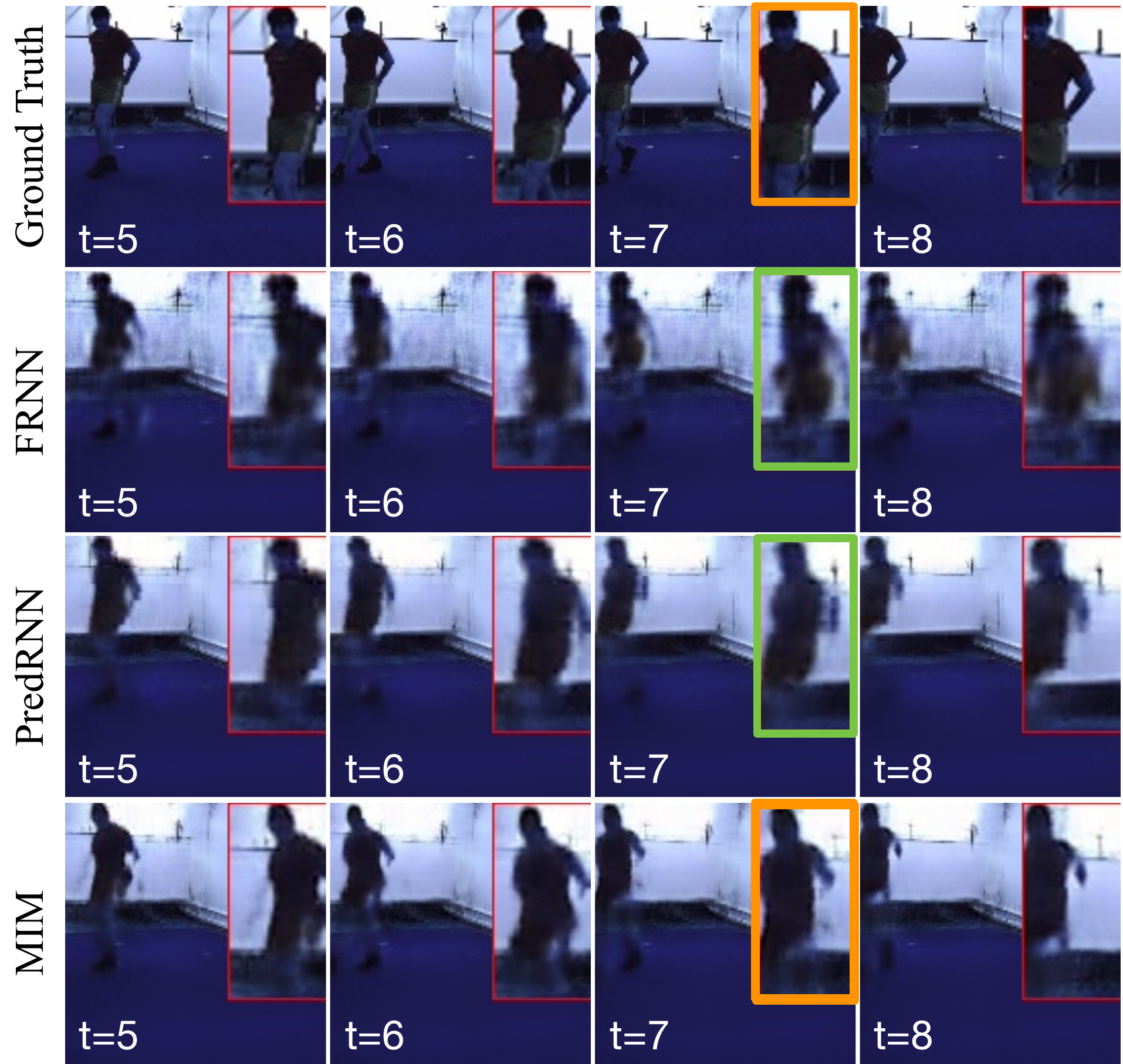}
    \vspace{-10pt}
    \caption{
    Examples of the generated images on the Human3.6M dataset. We zoom in to show the details of images.
    }
    \label{fig:human}
\end{figure}

As shown in Table \ref{tab:human} and Figure \ref{fig:human}, the MIM network outperforms the previous state-of-the-art models in both numerical metrics and visual effects.
The generated frames by MIM are more accurate in motion positions, as indicated by the orange boxes (versus the green boxes).
We notice that some work performs well on human-body datasets by modeling the structures of human joints, while the MIM network is designed for non-structural, general-purpose spacetime data such as RGB videos.
The structural and non-structural methods can be integrated and jointly used in a complementary manner, which is left for future work.

\begin{table}[h]
    \begin{center}
    \begin{tabular}{|l|ccc|}
    \hline
    Model & SSIM & MSE & MAE\\
    \hline\hline
    FRNN \citep{oliu2017folded} & 0.771 & 497.7 & 1901.1 \\
    PredRNN \citep{wang2017predrnn} & 0.781 & 484.1 & 1895.2 \\
    \textbf{MIM} & \textbf{0.790} & \textbf{429.9} & \textbf{1782.8} \\
    \hline
    \end{tabular}
    \end{center}
    \vspace{-5pt}
    \caption{
    Quantitative results on the Human3.6M dataset.
    }
    \label{tab:human}
\end{table}

\section{Conclusions}

We investigate the underlying non-stationarity that forms one of the main obstacles in spatiotemporal prediction. 
Existing LSTM-based models for spatiotemporal prediction are powerful in modeling difference-stationary sequences, whose capability for modeling high-order non-stationary process is limited by their relatively weak forget gates.
%
%
This paper proposes a new recurrent neural network to enable non-stationary modeling in the spacetime contexts. 
We formulate high-order non-stationarity as high-order polynomials with respect to the statistically varying trends of pixel intensities.
The order of non-stationary polynomials can be reduced by a couple of differencing operations. We leverage this idea from time-series analysis: stacking MIMs with differential inputs layer by layer.
The Memory In Memory (MIM) block is derived to model the complicated variations, which uses two cascaded recurrent modules to handle the non-stationary and approximately stationary components in the spatiotemporal dynamics. MIM achieves the state-of-the-art prediction performance on four datasets: a synthetic dataset of flying digits, a traffic flow prediction dataset, a weather forecasting dataset, and a human pose video dataset.

\section*{Acknowledgements} 
This work is supported by National Natural Science Foundation of China (No.~61772299, 71690231, and 61672313).

{\small
\bibliographystyle{ieee}
\bibliography{5165}
}

\end{document}